\documentclass[11pt,a4paper]{article}
\usepackage[hyperref]{naaclhlt2019}
\usepackage{times}
\usepackage{latexsym}
\usepackage{graphicx}
\usepackage{booktabs}
\usepackage{multirow}
\usepackage{xspace}
\newcommand{\eg}{\textit{e.g.}\xspace}
\newcommand{\ie}{\textit{i.e.}\xspace}
\newcommand{\ia}{\textit{i.a.}\xspace}
\newcommand{\F}{$\textrm{F}_1$\xspace}
\usepackage{url}
\aclfinalcopy
\def\aclpaperid{1358}

\title{Frowning Frodo, Wincing Leia, and a Seriously Great Friendship: 
Learning to Classify Emotional Relationships of Fictional Characters}

\author{
Evgeny Kim \and Roman Klinger\\
Institut f\"ur Maschinelle Sprachverarbeitung\\
University of Stuttgart\\
Pfaffenwaldring 5b, 70569 Stuttgart, Germany\\
\texttt{\{evgeny.kim,roman.klinger\}@ims.uni-stuttgart.de}
}

\date{}

\begin{document}
\maketitle
\begin{abstract}
  The development of a fictional plot is centered around characters
  who closely interact with each other forming dynamic social
  networks. In literature analysis, such networks have mostly been
  analyzed without particular relation types or focusing on roles
  which the characters take with respect to each other. We argue that
  an important aspect for the analysis of stories and their
  development is the emotion between characters.  In this paper, we
  combine these aspects into a unified framework to classify emotional
  relationships of fictional characters. We formalize it as a new task
  and describe the annotation of a corpus, based on fan-fiction short
  stories. The extraction pipeline which we propose consists of
  character identification (which we treat as given by an oracle here)
  and the relation classification. For the latter, we provide results
  using several approaches previously proposed for relation
  identification with neural methods. The best result of 0.45 \F is
  achieved with a GRU with character position indicators on the task
  of predicting undirected emotion relations in the associated social
  network graph.
\end{abstract}

\section{Introduction}
\begin{figure}[b]
  \centering
  \includegraphics{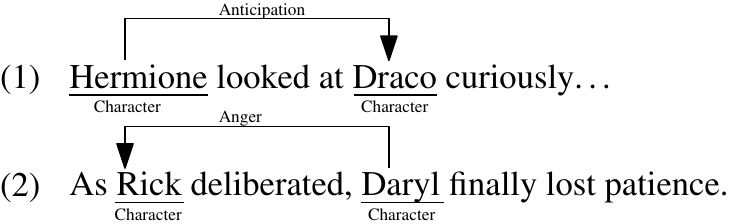}
  \caption{Examples for Emotional Character Interaction. (1) taken
    from \newcite{zephyr}, (2) from \newcite{emmyr}. The arrow starts
    at the experiencer and points at the causing character.}
  \label{fig:example}
\end{figure}

Every fictional story is centered around characters in conflict
\cite{ingermanson2009writing} which interact, grow closer or apart, as
each of them has ambitions and concrete goals
\cite[~p. 9]{emothesaurus}. Previous work on computational literary
studies includes two tasks, namely social network analysis and
sentiment/emotion analysis, both contributing to a computational
understanding of narrative structures. We argue that joining these two
tasks leverages simplifications that each approach makes when
considered independently.  We are not aware of any such attempt and
therefore propose the task of emotional character network extraction
from fictional texts, in which, given a text, a network is to be
generated, whose nodes correspond to characters and edges to emotions
between characters. One of the characters is part of a trigger/cause
for the emotion experienced by the other. Figure~\ref{fig:example}
depicts two examples for emotional character interactions at the text
level. Such relation extraction is the basis for generating social
networks of emotional interactions.

Dynamic social networks of characters are analyzed in previous work
with different goals, \eg, to test the differences in interactions
between various adaptations of a book \cite{agarwal2013automatic}; to
understand the correlation between dialogue and setting
\cite{elson2010extracting}; to test whether social networks derived
from Shakespeare's plays can be explained by a general sociological
model \cite{nalisnick2013extracting}; in the task of narrative
generation \cite{sack2013narrative}; to better understand the nature
of character interactions \cite{piper2017networks}. Further, previous
work analyses personality traits of characters (mostly) independently
of each other
\cite{massey2015annotating,Barth2018,bamman2014bayesian}.

Emotion analysis in literature has focused on the development of
emotions over time, abstracting away who experiences an emotion
\cite[\ia]{reagan2016emotional,elsner2015abstract,Kim2017investigating,Piper2015}. Fewer
works have addressed the annotation of emotion causes, \eg,
\newcite{neviarouskaya2013extracting}, \newcite{ghazi2015detecting},
\newcite{sauri2009factbank}, and \newcite{Kim2018}. To the best of our
knowledge, there is no previous research that deals with emotional
relationships of literary characters. The works that are conceptually
the closest to our paper are \newcite{chaturvedi2017unsupervised} and
\newcite{massey2015annotating}, who use a more general set of
relationship categories.

Most approaches to emotion classification from text build on the
classes proposed by \newcite{Plutchik2001} and
\newcite{ekman1992argument}. Here, we use a discrete emotion
categorization scheme based on fundamental emotions as proposed by
Plutchik. This model has previously been used in computational analysis of
literature \cite[\ia]{mohammad2012once}. We refer the reader to social
psychology literature for more details on the emotional relationship between people 
\cite{burkitt1997social,gaelick1985emotional}.

The main contributions of this paper are (1) to propose the new task
of emotional relationship classification of fictional characters, (2)
to provide a fan-fiction short story corpus annotated with characters
and their emotional relationships, and (3) to provide results for
relation extraction models for the task.
We evaluate our models on the textual and the social network graph
level and show that a neural model with positional indicators for
character roles performs the best. An additional analysis shows that the
task of character relationship detection leads to higher performance
scores for polarity detection than for more fine-grained emotion
classes. Differences between models are minimal when the task is cast
as a polarity classification but are striking for emotion classification.

This work has potential to support a literary scholar in analyzing
differences and commonalities across texts. As an example, one may
consider Goethe's \textit{The Sorrows of Young Werther}
\cite{werther1774}, a book that gave rise to a plethora of imitations
by other writers, who attempted to depict a similar love triangle
between main characters found in the original book. The results of our
study can potentially be used to compare the derivative works with the
original \cite[see also][]{Barth2018}.

\section{Corpus}
\paragraph{Data Collection and Annotation.}
Each emotion relation is characterized by a triple
$(C_{\textrm{exp}},e,C_{\textrm{cause}})$, in which the character
$C_{\textrm{exp}}$ feels the emotion $e$ (mentioned in text explicitly
or implicitly). The character $C_{\textrm{cause}}$ is part of an
event which triggers the emotion $e$. We consider 
the eight fundamental emotions defined by \newcite{Plutchik2001}
(anger, fear, joy, anticipation, trust, surprise, disgust,
sadness). Each character corresponds to a token sequence for the
relation extraction task and to a normalized entity in the graph
depiction.

Using WebAnno \cite{yimam2013webanno}, we annotate a sample of 19
complete English fan-fiction short stories, retrieved from the Archive
of Our Own project\footnote{\url{https://archiveofourown.org}} (due to
availability, the legal possibility to process the texts and a modern
language), and a single short story by \newcite{JoyceJ}
(Counterparts) being an exception from this genre in our corpus.  All
fan-fiction stories were marked by the respective author as complete,
are shorter than 1500 words, and depict at least four different
characters. They are tagged with the keywords ``emotion'' and
``relationships''.

The annotators were instructed to mark every character mention with a
canonical name and to decide if there is an emotional relationship
between the character and another character. If so, they marked the
corresponding emotion phrase with the emotion labels (as well as
indicating if the emotion is amplified, downtoned or negated). Based
on this phrase annotation, they marked two relations: from the emotion
phrase to the experiencing character and from the emotion phrase to
the causing character (if available, \ie, $C_{\textrm{cause}}$ can be
empty). One character may be described as experiencing multiple
emotions.

We generate a ``consensus'' annotation by keeping all emotion labels
by all annotators. This is motivated by the finding by
\newcite{Schuff2017} that such high-recall aggregation is better
modelled in an emotion prediction task.
The data is available at
\url{http://www.ims.uni-stuttgart.de/data/relationalemotions}.

\paragraph{Inter-Annotator Agreement}
We calculate the agreement along two dimensions, namely unlabelled
vs.\ labeled and instance vs.\ graph-level. Table \ref{iaafanfic}
reports the pairwise results for three annotators. In the
\emph{Inst.\ labelled} setting, we accept an instance being labeled as
true positive if both annotators marked the same characters as
experiencer and cause of an emotion and classified their interaction
with the same emotion. In the \emph{Inst.\ unlabelled} case, the
emotion label is allowed to be different. On the graph level
(\emph{Graph labelled} and \emph{Graph unlabelled}), the evaluation is
performed on an aggregated graph of interacting characters, \ie, a
relation is accepted by one annotator if the other annotator marked
the same interaction somewhere in the text. We use the F$_1$ score to
be able to measure the agreement between two annotators on the span
levels. For that, we treat the annotations from one annotator in the
pair as correct and the annotations from the other as predicted.

As Table \ref{iaafanfic} shows, agreement on the textual level is the
lowest with values between 19 and 33\,\% (depending on the annotator
pair), which also motivated our aggregation strategy mentioned
before. The values for graph-labelled agreement are more relevant for
our use-case of network generation. The values are higher
(66--93\,\%), showing that annotators agree when it comes to detecting
relationships regardless of where exactly in the text they appear.

\begin{table}[tbp]
\centering
\begin{tabular}{lccc} 
\toprule
& \multicolumn{1}{l}{a1--a2} & \multicolumn{1}{l}{a1--a3} & \multicolumn{1}{l}{a2--a3} \\
\cmidrule(lr){2-2}\cmidrule(lr){3-3}\cmidrule(lr){4-4}
Inst.\ labelled & 24 & 19 & 24 \\
Inst.\ unlab. & 33 & 27 & 29 \\
\cmidrule(lr){1-1}\cmidrule(lr){2-2}\cmidrule(lr){3-3}\cmidrule(lr){4-4}
Graph labelled & 66 & 69 & 66 \\
Graph unlabelled & 90 & 93 & 92 \\
\bottomrule
\end{tabular}%
\caption{F$_1$ scores in \,\% for agreement between annotators on different levels. a1, a2, and a3 are different annotators.}
\label{iaafanfic}%
\end{table}

\paragraph{Statistics.}
Table \ref{emorelations} summarizes the aggregated results of the
annotation. The column ``All" lists the number of experiencer
annotations (with an emotion), the column
``Rel."\ refers to the counts of emotion annotations with both
experiencer and cause.

\textit{Joy} has the highest number of annotated instances and the
highest number of relationship instances (413 and 308
respectively). In contrast, \textit{sadness} has the lowest number of
annotations with a total count of instances and relations being 97 and
64 respectively. Overall, we obtain 1335 annotated instances, which we
use to build and test our models.

\begin{table}[tbp]
  \centering
  \begin{tabular}{lrrrr}
    \toprule
    Emotion & & All & & Rel. \\ 
    \cmidrule{1-1}\cmidrule{3-3}\cmidrule{5-5}
    anger &  & 258 & & 197 \\
    anticipation &  & 307 & & 239 \\
    disgust &  & 163 & & 122 \\
    fear &  & 182 & &  120 \\
    joy & & 413 & & 308 \\
    sadness & & 97 & & 64 \\
    surprise &  & 143 & & 129 \\
    trust & & 179 & & 156 \\
    \cmidrule{1-1}\cmidrule{3-3}\cmidrule{5-5}
    \textbf{total} &  & \textbf{1742} & & \textbf{1335} \\
    \bottomrule
  \end{tabular}
  \caption{Statistics of emotion and relation annotation. ``All''
    indicates the total number of emotion annotations. ``Rel.''
    indicates total number of emotional relationships (including a
    causing character) instantiated with the given emotion.}
  \label{emorelations}
\end{table}

\section{Methods}
\begin{figure*}[t]
  \includegraphics[width=\textwidth]{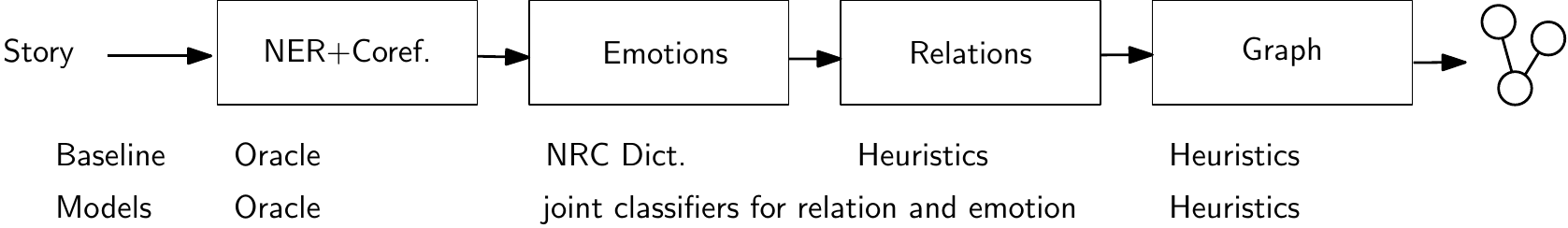}
  \caption{Models for the emotional relationship prediction. Oracle: a
    set of character pairs from the gold data.}
  \label{fig:baselines}
\end{figure*}
\begin{table}[t]
  \centering
  \setlength\tabcolsep{1pt} 
  \begin{tabular}{llll}
    \toprule
    Indicator & & Implementation example  \\ 
    \cmidrule{1-1}\cmidrule{3-3}
    No-Ind. & & Alice is angry with Bob  \\
    Role & & \texttt{<e>}Alice\texttt{</e>}\ldots\texttt{<c>}Bob\texttt{</c>} \\
    MRole & & \texttt{<e>}\ldots\texttt{<c>} \\
    Entity & & \texttt{<et>}Alice\texttt{</et>}\ldots\texttt{<et>}Bob\texttt{</et>}\\
    MEntity & & \texttt{<et>}\ldots\texttt{</et>}\\
    \bottomrule
  \end{tabular}
  \caption{Different indicators applied to the same instance. \textit{No-Ind.} means no positional indicators are added. \textit{M} in \textit{MRole} and \textit{MEntity} means that the name of the character is masked. Tag \texttt{<e>} indicates the experiencer. Tag \texttt{<c>} indicates the cause. Tag \texttt{<et>} indicates an entity.}
  \label{indicators}
\end{table}
Figure \ref{fig:baselines} depicts the process flow for each of the
models. We distinguish between \textit{directed} and
\textit{undirected} relation prediction. In the directed scenario, we
classify which character is the experiencer and which character is the
cause, as well as what is the emotion between two characters. For the
undirected scenario, we only classify the emotion relation between two
characters. We do not tackle character name recognition here: our
models build on top of gold character annotations.

The \emph{baseline} model predicts the emotion for a character pair
based on the NRC dictionary \cite{Mohammad2013}. It accepts the
emotion associated with the words occurring in a window of $n$
tokens around the two characters, with $n$ being a parameter
set based on results on a development set for each model (see
supplementary material for more details).

Further we cast the relation detection as a machine learning-based
classification task, in which each classification instance consists of
two character mentions with up to $n$ tokens context to the left and
to the right of the character mentions. We compare an extremely randomized
tree classifier with bag-of-words features \cite{geurts2006extremely}
(\emph{BOW-RF}) with a two-layer GRU neural network
\cite{chung2014empirical} with max and averaged pooling. In the
latter, we use different variations of encoding the character
positions with indicators (inspired by \newcite{zhou2016attention},
who propose the use of positional indicators for relation
detection). Our variations are exemplified in Table
\ref{indicators}. Note that the case of predicting directed relations
is simplified in the ``Role'' and ``MRole'' cases in contrast to
``Entity'' and ``MEntity'', as the model has access to gold
information about the relation direction.

We obtain word vectors for the embedding layer from GloVe \cite[pre-trained
on Common Crawl, $d=300$, ][]{pennington2014glove} and initialize
out-of-vocabulary terms with zeros (including the position
indicators).

\section{Experiments}
\begin{table}[t]
\centering
\setlength\tabcolsep{3pt} 
  \begin{tabular}{llrrrrrr}
    \toprule
    && \multicolumn{3}{c}{Undir.} & \multicolumn{3}{c}{Directed}\\
    \cmidrule(l){3-5}\cmidrule(l){6-8}
    &Model & 8c & 5c & 2c& 8c& 5c& 2c \\
    \cmidrule(l{1pt}r{1pt}){2-2}\cmidrule(l{1pt}r{1pt}){3-3}\cmidrule(l{1pt}r{1pt}){4-4}\cmidrule(l{1pt}r{1pt}){5-5}\cmidrule(l{1pt}r{1pt}){6-6}\cmidrule(l{1pt}r{1pt}){7-7}\cmidrule(l{1pt}r{1pt}){8-8}
    \multirow{7}{*}{\rotatebox{90}{Dev}}
    \multirow{7}{*}{\rotatebox{90}{Instance level}}
    &Baseline  & 24 & 30 &  56 & -- & -- & --  \\
    &BOW-RF  & 18 & 31 & 56 & 20 & 19 & 35  \\
    &GRU+NoInd. & 31 & 39 & 64 &26 & 23& 37   \\
    &GRU+Role & 19 & 35 &  55& 33 & 34 & 57 \\
    &GRU+MRole  & 30 & 44 & 67 &38 & 44 & 65 \\
    &GRU+Entity & 20 & 34 & 58 & 23 & 19 & 30\\
    &GRU+MEntity & 30 & 43 & 65 & 28 & 29 & 40 \\
\cmidrule(l{1pt}r{1pt}){1-2}\cmidrule(l{1pt}r{1pt}){3-3}\cmidrule(l{1pt}r{1pt}){4-4}\cmidrule(l{1pt}r{1pt}){5-5}\cmidrule(l{1pt}r{1pt}){6-6}\cmidrule(l{1pt}r{1pt}){7-7}\cmidrule(l{1pt}r{1pt}){8-8}
    \multirow{7}{*}{\rotatebox{90}{Dev}}
    \multirow{7}{*}{\rotatebox{90}{Story level}}
    &Baseline & 24 & 31 & 56 & -- & -- & --\\
    &BOW-RF   & 21 & 35 & 58 & 22 & 20 & 38\\
    &GRU+NoInd. & 33 & 41 & 66& 25 & 23& 38 \\
    &GRU+Role & 19 & 34& 55& 33 & 35 & 56\\
    &GRU+MRole & 32& 44 & 67 &39 & 44& 65\\
    &GRU+Entity & 21 & 31 & 57 & 22 & 18 & 30\\
    &GRU+MEntity & 33 & 46 & 65 & 28 & 30 & 39\\
\cmidrule(l{1pt}r{1pt}){1-2}\cmidrule(l{1pt}r{1pt}){3-3}\cmidrule(l{1pt}r{1pt}){4-4}\cmidrule(l{1pt}r{1pt}){5-5}\cmidrule(l{1pt}r{1pt}){6-6}\cmidrule(l{1pt}r{1pt}){7-7}\cmidrule(l{1pt}r{1pt}){8-8}
    \multirow{7}{*}{\rotatebox{90}{Dev}}
    \multirow{7}{*}{\rotatebox{90}{Graph-level}} 
    &Baseline & 31 & 46 & 65  & -- & -- & --\\
    &BOW-RF & 27 & 36 & 71 & 34  & 34 & 54\\
    &GRU+NoInd. & 44 & 55 &  73 & 35 & 33 & 54 \\
    &GRU+Role & 35 & 49 & 65 & 41 & 43 & 57 \\
    &GRU+MRole & 45 & 58 & 73 & 40 & 48 & 65 \\
    &GRU+Entity & 37 & 50 & 68 & 39 & 29 & 49\\    
    &GRU+MEntity & 47 & 63 & 73 & 39 & 39 & 52 \\
\cmidrule(l{1pt}r{1pt}){1-2}\cmidrule(l{1pt}r{1pt}){3-3}\cmidrule(l{1pt}r{1pt}){4-4}\cmidrule(l{1pt}r{1pt}){5-5}\cmidrule(l{1pt}r{1pt}){6-6}\cmidrule(l{1pt}r{1pt}){7-7}\cmidrule(l{1pt}r{1pt}){8-8}
    \multirow{3}{*}{\rotatebox{90}{Test}}
  &\hspace{-2mm}GRU+MRole Inst. & 30 & 44 & 64  & 38 & 43 & 65 \\
  &\hspace{-2mm}GRU+MRole Story & 33 & 45 & 65 & 39 & 43 & 66 \\
  &\hspace{-2mm}GRU+MRole Graph & 45 & 59 & 71 & 42 & 49 & 66 \\
    \bottomrule
  \end{tabular}
  \caption{Cross-validated results in \% F$_1$ score, average of four runs. Inst.\ level: aggregated over all instances in the
    dataset. Story level: averaged performance on all
    stories. Graph-level: averaged performance on graph level on all
    stories. Test results are reported for the best indicator type. See Table \ref{indicators} for the examples of the indicator implementation.}
  \label{dev}
\end{table}

\paragraph{Experimental Setting.}
In the classification experiments, we compare the performance of our
models on different label sets. Namely, we compare the complete
emotion set with 8 classes to a 5 class scenario where we join
\textit{anger} and \textit{disgust}, \textit{trust} and \textit{joy},
as well as \textit{anticipation} and \textit{surprise} (based on
preliminary experiments and inspection of confusion matrices). The
2-class scenario consists of positive (\textit{anticipation, joy,
  trust, surprise}) and negative relations (\textit{anger, fear,
  sadness, disgust}). For each set of classes, we consider a setting
where directed relations are predicted with one where the direction is
ignored.  Therefore, in the \textit{directed} prediction scenario,
each emotion constitutes two classes to be predicted for both possible
directions (therefore, 16, 10, and 4 labels exist).

The evaluation is performed with precision, recall and \F in a
cross-story validation setting, in which each story is used as one
separate test/validation source.  For model selection and
meta-parameter optimization, we use 50\,\% randomly sampled
annotations from this respective test/validation instance as a
validation set and the remainder as test data.

Further, we evaluate on three different levels of granularity: Given
two character mentions, in the instance-level evaluation, we only
accept the prediction to be correct if exactly the same mention has
the according emotion annotation. We then aggregate the different true
positive, false positive and false negative values across all stories
before averaging to an aggregated score (similar to micro-averaging).
On the story-level, we also accept a prediction to be a true positive
the same way, but first calculate the result P/R/\F for the whole
story before averaging (similar to macro-averaging).  On the
graph-level, we accept a prediction for a character pair to be correct
without considering the exact position.

\paragraph{Results.}
Table~\ref{dev} shows the results (precision and recall shown in
supplementary \ material) on development data and independent test
data for the best models. The GRU+MRole model achieves the highest
performance with improvement over BOW-RF on the instance and story
levels, and shows a clear improvement over the GRU+NoInd. model in the
directed 8-class setting. GRU+Role achieves the highest performance on
the graph level in the directed 8-class setting. In the undirected
prediction setting, all models perform better in the 5-class
experiment and 2-class experiment than in 8-class experiment. This is
not always the case for the directed prediction, where some models
perform better in 8-class experiment (GRU+NoInd., GRU+Entity, BOW-RF).

We observe that the difference in F$_1$ score between the baseline,
bag-of-words model and our GRU models in a 2-class experiment is
marginal. This may be an indicator that the binary representation
harms the classification of emotional relations between characters, as
they can be nuanced and do not always perfectly map to either positive
and negative classes. On the other side, a more sophisticated
classification approach is necessary to capture these nuanced
differences.

As expected, we observe a better performance on a graph level for all
models, with the highest performance of 47\,\% \F (GRU+MEntity),
63\,\% \F (GRU+MEntity), and 73\,\% \F (GRU+MRole, GRU+MEntity,
GRU+NoInd.) in undirected 8-, 5-, and 2-class experiments,
respectively, on the development set. In the directed scenario, the
highest performances are 41\,\% \F (GRU+Role), 48\,\% \F (GRU+MRole),
and 65\,\% \F (GRU+MRole).

The results show that the sequential and embedding
information captured by a GRU as well as additional positional
information are all relevant for a substantial performance, at least
on the fine-grained emotion prediction task.

\section{Conclusion \& Future Work}
In this paper, we formulated the new task of emotional character
network extraction from fictional texts. We argued that joining social
network analysis of fiction with emotion analysis leverages
simplifications that each approach makes when considered
independently. We presented a publicly available corpus of fan-fiction
short stories annotated with character relations and proposed
several relation classification models. We showed that a recurrent
neural architecture with positional indicators leads to the best
results of relation classification. We also showed that
differences between different machine learning models with binary
mapping of emotion relation is almost leveled. This may suggest
that emotion relation classification is best modeled in a
multi-class setting, as emotional interactions of fictional characters
are nuanced and do not simply map to either a positive or a negative
class.

For future work we propose to develop a real-world application
pipeline in which character pairs are not given by an oracle, but
rather extracted from text automatically using named entity
recognition. To better understand the relation between instance and
graph levels, we propose to explore the best strategy for edge
labeling either by a majority vote or accepting the edges with the
highest confidence scores. Further, modeling the task in an end-to-end
learning setting from text to directly predict the graph, in the
spirit of multi-instance learning, is one of the next steps. To that
end, we suggest obtaining more gold data with character relations
and optimize the pipeline towards the best performance on additional
data.

\section*{Acknowledgements}
This research has been conducted within the CRETA project
(\url{http://www.creta.uni-stuttgart.de/}) which is funded by the
German Ministry for Education and Research (BMBF) and partially funded
by the German Research Council (DFG), projects SEAT (Structured
Multi-Domain Emotion Analysis from Text, KL 2869/1-1). We thank
Laura-Ana-Maria Bostan and Heike Adel for fruitful discussions.

\onecolumn
\appendix

\section{Supplementary Material}

\subsection{Complete Result Table}
Table \ref{tab:results} contains the complete results with precision,
recall and \F.
\vspace{1cm}
\begingroup

  \centering
  \small
  \setlength{\tabcolsep}{3pt}
  \begin{tabular}{llrrrrrrrrrrrrrrrrrr}
    \toprule
    && \multicolumn{9}{c}{\textbf{Undirected}} & \multicolumn{9}{c}{\textbf{Directed}}\\
    \cmidrule(l){3-11}\cmidrule(l){12-20}
    && \multicolumn{3}{c}{8 Class} & \multicolumn{3}{c}{5 Class} & \multicolumn{3}{c}{2 Class} & \multicolumn{3}{c}{8 Class} & \multicolumn{3}{c}{5 Class} & \multicolumn{3}{c}{2 Class} \\
    \cmidrule(l){3-5}\cmidrule(lr){6-8}\cmidrule(lr){9-11}\cmidrule(l){12-14}\cmidrule(lr){15-17}\cmidrule(lr){18-20}
    && P & R & F$_1$ & P & R & F$_1$ & P & R & F$_1$ & P & R & F$_1$ & P & R & F$_1$ & P & R & F$_1$ \\ 
\cmidrule(l{1pt}r{1pt}){3-3}\cmidrule(l{1pt}r{1pt}){4-4}\cmidrule(l{1pt}r{1pt}){5-5}\cmidrule(l{1pt}r{1pt}){6-6}\cmidrule(l{1pt}r{1pt}){7-7}\cmidrule(l{1pt}r{1pt}){8-8}\cmidrule(l{1pt}r{1pt}){9-9}\cmidrule(l{1pt}r{1pt}){10-10}\cmidrule(l{1pt}){11-11}\cmidrule(l{1pt}){12-12}\cmidrule(l{1pt}){13-13}\cmidrule(l{1pt}){14-14}\cmidrule(l{1pt}){15-15}\cmidrule(l{1pt}){16-16}\cmidrule(l{1pt}){17-17}\cmidrule(l{1pt}){18-18}\cmidrule(l{1pt}){19-19}\cmidrule(l{1pt}){20-20}
    \multirow{7}{*}{\rotatebox{90}{Dev}}
    \multirow{7}{*}{\rotatebox{90}{Instance level}}
    &Baseline & 19 &31&24 &25&38 & 30&39&100 & 56 &  & & & & & & & &  \\
    &BOW-RF  & 18& 18& 18& 31& 31& 31&56& 56& 56 & 20& 20& 20& 19& 19& 19&35 &35 &35 \\
    &GRU+NoInd. & 31 & 31 & 31 & 39 & 39 &39 &64 &64 &64 &26 &26 & 26& 23& 23& 23& 37& 37& 37  \\
    &GRU+Role & 19& 19&19 &35 &35 &35 & 55& 55& 55& 33& 33& 33& 34& 34& 34& 57& 57& 57  \\
    &GRU+MaskRole &30 & 30 & 30& 44&44 &44 &67 & 67& 67&38 &38 &38 &44 &44 &44 &65 & 65& 65\\
    &GRU+Entity &20 &20 &20 &34 &34 &34 &58 &58 &58 &23 &23 &23 &19 &19 &19 &30 &30 &30 \\
    &GRU+MaskEntity &  30& 30 &30  & 43 &43  &43  &65  &65  & 65 &28  &28  & 28 & 29  & 29 & 29 & 40 &40  & 40 \\
\cmidrule(l{1pt}r{1pt}){1-2}\cmidrule(l{1pt}r{1pt}){3-3}\cmidrule(l{1pt}r{1pt}){4-4}\cmidrule(l{1pt}r{1pt}){5-5}\cmidrule(l{1pt}r{1pt}){6-6}\cmidrule(l{1pt}r{1pt}){7-7}\cmidrule(l{1pt}r{1pt}){8-8}\cmidrule(l{1pt}r{1pt}){9-9}\cmidrule(l{1pt}r{1pt}){10-10}\cmidrule(l{1pt}){11-11}\cmidrule(l{1pt}){12-12}\cmidrule(l{1pt}){13-13}\cmidrule(l{1pt}){14-14}\cmidrule(l{1pt}){15-15}\cmidrule(l{1pt}){16-16}\cmidrule(l{1pt}){17-17}\cmidrule(l{1pt}){18-18}\cmidrule(l{1pt}){19-19}\cmidrule(l{1pt}){20-20}
    \multirow{7}{*}{\rotatebox{90}{Dev}}
    \multirow{7}{*}{\rotatebox{90}{Story level}}
    &Baseline & 20 & 32 & 24& 27& 39& 31& 40& 100& 56& & & & & & & & &  \\
    &BOW-RF   & 20 & 24& 21& 33 &36 &35 & 58& 59& 58& 21& 25& 22&19 &23 &20 &37 &39 &38\\
    &GRU+NoInd. & 33 &33 &33 & 41& 41& 41& 66& 66&66 &25 &25 & 25& 23& 23& 23& 38& 38& 38\\
    &GRU+Role &19 &19 &19 &34 &34 &34 &55 &55 &55 &33 &33 &33 & 35& 35& 35&56 &56 &56 \\
    &GRU+MaskRole &32 &32 &32 &44 &44 &44 &67 &67 &67 &39 &39 &39 &44 &44 &44 &65 &65 &65\\
    &GRU+Entity &21 &21 &21 &31 &31 &31 &57 &57 & 57&22 & 22& 22& 18& 18& 18& 30& 30&30\\
    &GRU+MaskEntity & 33&33 &33 & 46& 46& 46& 65& 65& 65& 28& 28& 28& 30& 30& 30& 39& 39&39\\
\cmidrule(l{1pt}r{1pt}){1-2}\cmidrule(l{1pt}r{1pt}){3-3}\cmidrule(l{1pt}r{1pt}){4-4}\cmidrule(l{1pt}r{1pt}){5-5}\cmidrule(l{1pt}r{1pt}){6-6}\cmidrule(l{1pt}r{1pt}){7-7}\cmidrule(l{1pt}r{1pt}){8-8}\cmidrule(l{1pt}r{1pt}){9-9}\cmidrule(l{1pt}r{1pt}){10-10}\cmidrule(l{1pt}){11-11}\cmidrule(l{1pt}){12-12}\cmidrule(l{1pt}){13-13}\cmidrule(l{1pt}){14-14}\cmidrule(l{1pt}){15-15}\cmidrule(l{1pt}){16-16}\cmidrule(l{1pt}){17-17}\cmidrule(l{1pt}){18-18}\cmidrule(l{1pt}){19-19}\cmidrule(l{1pt}){20-20}
    \multirow{8}{*}{\rotatebox{90}{Dev}}
    \multirow{8}{*}{\rotatebox{90}{Graph-level}} 
    &Baseline &36 &38 &31 & 50& 41& 46& 88 &52 &65 & & & & & & & & & \\
    &BOW-RF & 68 &17 &27 & 72& 35& 36& 70& 72& 71& 72& 23& 34& 79& 23& 34&54 &54 &54\\
    &GRU+NoInd. & 44 &44 &44 &55 &55 &55 &73 & 73& 73& 35& 35& 35& 33&33 &33 &54 &54 &54 \\
    &GRU+Role & 35& 35& 35& 49& 49& 49& 65& 65& 65& 41& 41& 41& 43& 43& 43& 57& 57&57\\
    &GRU+MaskRole &45 &45 & 45& 58& 58& 58& 73& 73& 73& 40& 40& 40& 48& 48& 48& 65& 65&65\\
    &GRU+Entity &37 &37 &37 & 50&50 &50 &68 &68 &68 &39 &39 &39 &29 &29 &29 &49 &49 &49\\    
    &GRU+MaskEntity & 47& 47& 47& 63& 63& 63& 73& 73& 73& 39& 39& 39& 39& 39& 39& 52& 52&52\\
\cmidrule(l{1pt}r{1pt}){1-2}\cmidrule(l{1pt}r{1pt}){3-3}\cmidrule(l{1pt}r{1pt}){4-4}\cmidrule(l{1pt}r{1pt}){5-5}\cmidrule(l{1pt}r{1pt}){6-6}\cmidrule(l{1pt}r{1pt}){7-7}\cmidrule(l{1pt}r{1pt}){8-8}\cmidrule(l{1pt}r{1pt}){9-9}\cmidrule(l{1pt}r{1pt}){10-10}\cmidrule(l{1pt}){11-11}\cmidrule(l{1pt}){12-12}\cmidrule(l{1pt}){13-13}\cmidrule(l{1pt}){14-14}\cmidrule(l{1pt}){15-15}\cmidrule(l{1pt}){16-16}\cmidrule(l{1pt}){17-17}\cmidrule(l{1pt}){18-18}\cmidrule(l{1pt}){19-19}\cmidrule(l{1pt}){20-20}
    \multirow{3}{*}{\rotatebox{90}{Test}}
  & GRU+MaskRole Inst. & 30&30 &30 &44 &44 &44 &64 &64 & 64& 38& 38& 38& 43& 43& 43& 65& 65& 65\\
    & GRU+MaskRole Story &33 &33 &33 &45 &45 & 45& 65& 65& 65& 39& 39& 39& 43& 43& 43& 66& 66&66\\
  & GRU+MaskRole Graph &45 &45 &45 &59 &59 &59 &71 &71 &71 &42 &42 & 42& 49& 49& 49& 66& 66&66\\
    \bottomrule
  \end{tabular}
  \captionof{table}{Cross-validated results for different models in percentages of F$_1$ score. Inst. level: aggregated over all instances in the dataset. Story level: averaged performance on all stories. Graph-level: averaged performance on graph level on all stories. Test results are reported for the best indicator type. \textit{GRU+NoInd.}: Alice is angry with Bob. \textit{GRU+Role}: \texttt{<exp>}Alice\texttt{</exp>}\ldots\texttt{<target>}Bob\texttt{</target>}. \textit{GRU+MaskRole}:\texttt{<exp>}\ldots\texttt{<target>}. \textit{GRU+Entity}: \texttt{<ent>}Alice\texttt{</ent>}\ldots\texttt{<char>}Bob\texttt{</char>}. \textit{GRU+MaskEntity}: \texttt{<ent>}\ldots\texttt{</ent>}.}
  \label{tab:results}

\endgroup

\clearpage

\subsection{Results as Plots}
\includegraphics[page=1, width=\linewidth]{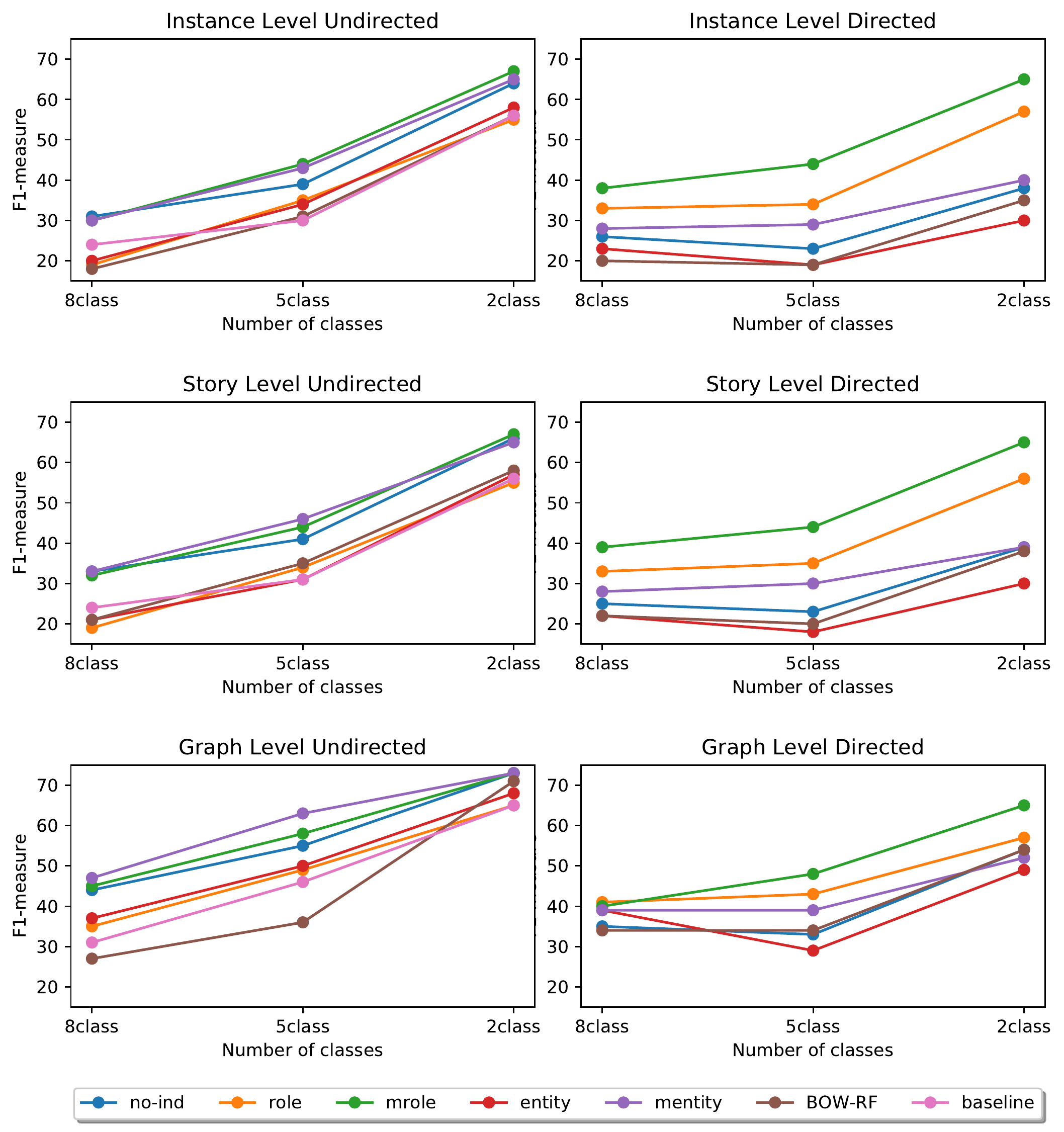}
The plots show the performance of our models with different number of
modeled classes. One may observe that all models perform better in a
2-class scenario (directed and undirected). However, the differences
between the models in a 2-class setting are marginal, especially in
the undirected scenario. This may suggest that character relations are
more nuanced than binary. It also suggests that directionality is an
important aspect for the task of relation classification. In the
directed classification scenario, the differences between different
models are more pronounced, as compared to the undirected scenario.
\clearpage
\subsection{Window Size Experiments on Instance-level}
\includegraphics[page=1, width=\linewidth]{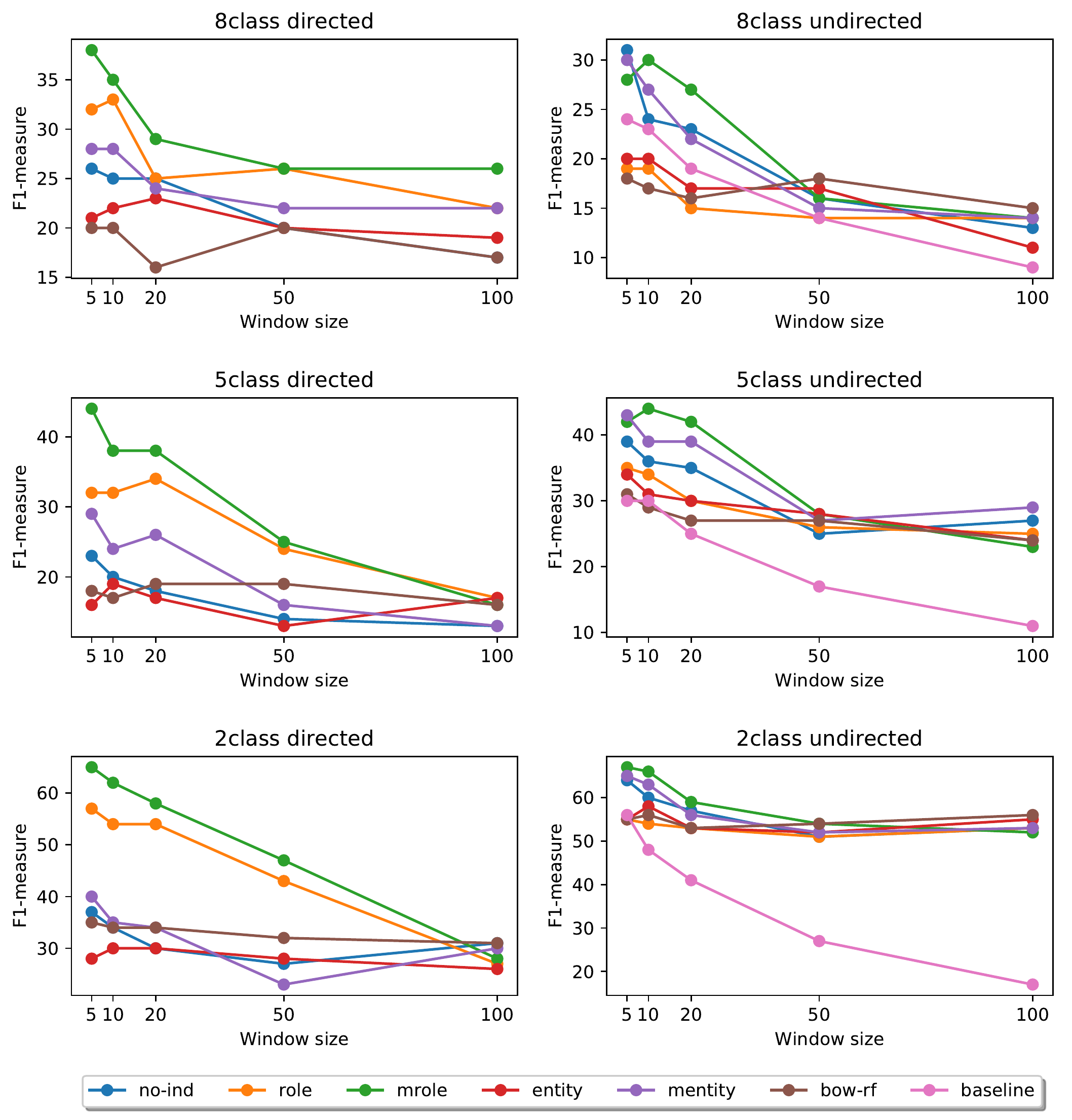}
The plots depict the performance of all models evaluated on the
instance-level for one example run. We tuned the window size parameter on a development
set using a set of window sizes of 5, 10, 20, 50, and 100 tokens
around character mentions. As one may see, the window size of 5 tokens
is the best in the majority of cases. The GRU+Entity model shows an
exception as it achieves the highest performance with 20 tokens in the
8-class directed scenario. The 2-class GRU+Entity works best with 10
tokens around the character mentions.

\end{document}